\documentclass[11pt]{article}

\usepackage[final]{acl}

\usepackage{pslatex} 
\usepackage{latexsym}
\usepackage{amsmath}
\usepackage{booktabs}
\usepackage{tabularx}
\usepackage{enumitem}
\usepackage{amssymb}
\usepackage{pifont}
\usepackage{comment}

\usepackage[T1]{fontenc}

\usepackage[utf8]{inputenc}

\usepackage{microtype}
\usepackage{float}
\usepackage{inconsolata}

\usepackage{graphicx}

\usepackage{tikz}
\usetikzlibrary{positioning, calc, arrows.meta, shapes.geometric}

\usepackage{makecell}
\usepackage{xcolor}
\usepackage{multirow}

\usepackage[colorinlistoftodos,prependcaption,textsize=scriptsize]{todonotes}
\definecolor{wrong}{RGB}{248, 0, 0}
\definecolor{correct}{RGB}{26, 148, 49}

\title{Think Before you Write:\\
  QA-Guided Reasoning for Character Descriptions in Books}

\author{Argyrios Papoudakis \qquad Mirella Lapata \qquad Frank Keller \\
Institute of Language, Cognition and Computation \\ 
School of Informatics, University of Edinburgh \\
10 Crichton Street, Edinburgh EH8 9AB \\
\texttt{a.papoudakis@sms.ed.ac.uk}, \texttt{\{mlap, keller\}@inf.ed.ac.uk}}

\begin{document}
\maketitle
\begin{abstract}

Character description generation is an important capability for
narrative-focused applications such as summarization, story analysis,
and character-driven simulations. However, generating accurate
character descriptions from long-form narratives (e.g., novels) is
challenging: models must track evolving attributes (e.g.,
relationships and events), integrate evidence scattered across the
text, and infer implicit details. Despite the success of
reasoning-enabled LLMs on many benchmarks, we find that for character
description generation their performance improves when built-in
reasoning is disabled (i.e., an empty reasoning trace). Motivated by
this, we propose a training framework that decouples reasoning from
generation. Our approach, which can be applied on top of long-context
LLMs or chunk-based methods, consists of a reasoning model that
produces a structured QA reasoning trace and a generation model that
conditions on this trace to produce the final character description.
Experiments on two datasets (BookWorm and CroSS) show that QA-guided
reasoning improves faithfulness, informativeness, and grounding over
strong long-context baselines\footnote{We release our code and data at \url{https://github.com/apapoudakis/qa-guided-reasoning}}.

\end{abstract}

\section{Introduction}

\begin{figure}[t]
\centering

\begin{tikzpicture}[
    node distance=0.8cm and 1.2cm,
    box/.style={rectangle, draw, rounded corners, minimum height=0.9cm, minimum width=2.2cm, align=center, font=\small},
    model/.style={box, fill=blue!20},
    gen/.style={box,fill=red!20},
    data/.style={box, fill=gray!15},
    output/.style={box, fill=green!15},
    arrow/.style={->, >=stealth, thick},
    label/.style={font=\scriptsize, align=center}
]

\node[data] (book) {Book\\[-1pt]\scriptsize (full text)};
\node[data, right=0.6cm of book] (char) {Character\\[-1pt]\scriptsize name};

\node[model, below=1.2cm of $(book)!0.5!(char)$] (qa) {QA
  Reasoning\\[-1pt]Model $R_{\phi}$};

\node[right=1.8cm of qa, align=left, font=\small] (trace) {
    \begin{tabular}{@{}l@{}}
    \textbf{QA Trace}\\[2pt]
    \scriptsize Q\textsubscript{1}: What is X's role?\\[-1pt]
    \scriptsize A\textsubscript{1}: Protagonist\\[1pt]
    \scriptsize Q\textsubscript{2}: Who is X's ally?\\[-1pt]
    \scriptsize A\textsubscript{2}: Character Y\\[1pt]
    \scriptsize \quad\vdots
    \end{tabular}
};

\node[gen, below=1cm of qa] (gen) {Generation\\[-1pt]Model $G_{\theta}$};

\node[yshift=.75cm, right=0.5cm of gen, font=\scriptsize, align=left, text=gray!50!black, fill=blue!8, rounded corners=3pt, inner sep=5pt] (think) {
    \texttt{<think>}\\
    QA trace\\
    \texttt{</think>}
};

\node[output, below=1.0cm of gen, anchor=north, xshift=1.5cm, font=\scriptsize, align=left, text=black, 
      fill=green!8, rounded corners=3pt, inner sep=5pt, text width=7cm] (desc) {
    \textbf{Character Description:}\\[2pt]
    \textit{X's adventures begin with her fateful jump down the rabbit hole, and the tale is an extended metaphor for the challenges she will face as she grows into an adult. She possesses unusual composure for a child, and she seems bright but makes many charming mistakes...}
};

\draw[arrow] (book.south) -- ++(0,-0.4) -| ([xshift=-0.3cm]qa.north);
\draw[arrow] (char.south) -- ++(0,-0.4) -| ([xshift=0.3cm]qa.north);
\draw[arrow] (qa.east) -- (trace.west);
\draw[arrow] (trace.south) |- ([yshift=0.15cm]gen.east);
\draw[arrow] (gen.south) -- ([xshift=-1.5cm]desc.north);

\draw[arrow, dashed] (book.south) -- ++(0,-0.4) -- ++(-0.9,0) |- ([yshift=-0.15cm]gen.west);

\node[left=0.05cm of qa, font=\scriptsize, text=blue!60, align=right] {GRPO\\training};
\node[left=0.05cm of gen, yshift=.2cm, font=\scriptsize, text=red!60, align=right] {SFT\\training};

\end{tikzpicture}

\caption{Overview of QA-based guided reasoning for character
  description generation. Given a book and character name, the
  \textit{reasoning model} generates a structured QA trace capturing
  salient character information. This trace is injected into the
  thinking tokens of the \textit{generation model}, which produces the
  final character description. The reasoning model can be trained with
  GRPO and the generation model with SFT.} 
\label{fig:pipeline}
\vspace{-10pt}
\end{figure}
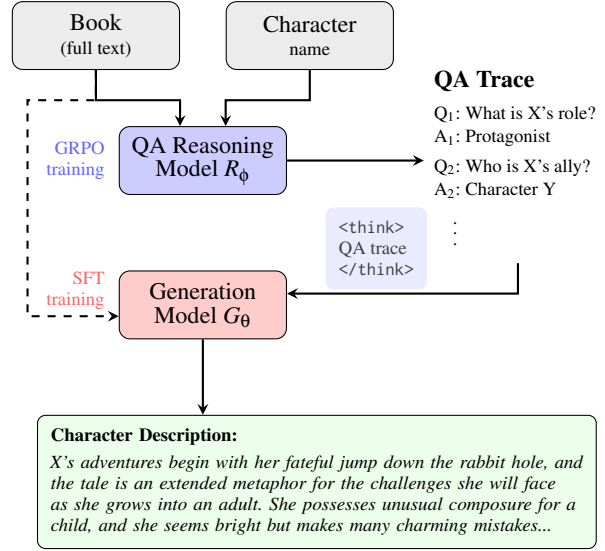

Writers craft characters to create engaging stories that invite
readers to experience actions, emotions, and goals from the
characters' perspectives. Characters form the core of the narrative,
with other elements such as plot, conflict, setting, and theme, built
around them. A growing body of work in natural language processing
aims to model characters for story analysis~\citep{zhu-etal-2023-nlp},
narrative generation and
summarization~\citep{fan-etal-2018-hierarchical}, and persona
simulation within character-driven
interactions~\citep{shao-etal-2023-character}. Yet despite their
importance, characters remain difficult to model computationally,
particularly in long-form narratives where their traits, motivations,
and relationships evolve over extended spans of text and through complex and often
non-linear plots~\citep{chaturvedi2015modeling, vishnubhotla-etal-2024-emotion}.

Consequently, a central difficulty in character-centric understanding stems from the sheer length of the input. To address this, earlier work adopted
chunk-based approaches using LLMs with short-context abilities to
process stories either hierarchically~\citep{wu2021recursively} or
incrementally~\citep{chang2023booookscore}. While these methods are
computationally efficient, they often struggle to capture all relevant
information within individual chunks and to integrate evidence
reliably across chunks.  Retrieval-augmented methods have also been
proposed to select salient context~\citep{xu2023retrieval}; however,
the retrieved passages are frequently disjoint and may omit critical
details, undermining global narrative coherence.  More recently,
long-context LLMs, capable of processing up to millions of
tokens~\cite{team2024gemini}, have been used to analyze stories
in a single pass. Nevertheless, even these models face known
limitations, including position bias~\citep{liu-etal-2024-lost},
difficulty in exploiting in-context examples~\citep{li2024long} and
challenges in integrating information across distant
spans~\citep{tian-etal-2025-distance}.

Beyond context scaling, long narratives present an additional
challenge: much information is implied rather than explicitly stated.
Models must therefore reason over long inputs (>100k tokens),
integrating evidence from different text passages. This requires
inferring implicit details, combining information which is being
revealed piecemeal, and filtering out irrelevant content.  Recent work
has enabled explicit reasoning via post-training reinforcement
learning~\citep{shao2024deepseekmath}, achieving
strong results on tasks such as mathematics \citep{yang2024qwen2} and
short-form question answering \citep{rein2024gpqa}. However, extending
these approaches to non-verifiable settings like text generation
remains challenging, as reward computation is difficult when many
valid outputs exist across multiple evaluation dimensions.

We address these limitations by focusing on character description
generation in books \citep{brahman-etal-2021-characters-tell}, where
models must produce factual descriptions given a character and the
full story. We propose a two-stage training framework that decouples
reasoning from generation: (1)~a \textit{reasoning model} produces an
explicit reasoning trace, and (2)~a \textit{generation model}
conditions on this trace to produce the final description (see
Figure~\ref{fig:pipeline}). Importantly, our reasoning model uses
question-answer pairs --- which have been previously shown to provide a useful
abstraction for isolating salient content in summarization
\cite{huot-etal-2023-text,narayan-etal-2023-conditional,10.1162/TACL.a.30}
--- as a scaffold to \emph{guide} downstream description generation,
forcing the model to ``reason through'' salient evidence \emph{before}
writing. As our reasoning model is input agnostic, it can be
integrated on top of different long-context methods, including
hierarchical merging, incremental updating, and direct processing with
long-context LLMs.

To train the reasoning model, we use Group Relative Policy
Optimization (GRPO)~\citep{shao2024deepseekmath} directly on
\emph{gold-standard} character descriptions and \emph{silver-standard}
reasoning traces derived from them, thereby avoiding the need to
simulate rewards based on the final generated outputs. 
The generation model can either be trained with supervised fine-tuning (SFT) on these reasoning traces or applied in zero-shot mode.

We evaluate our approach on two character understanding datasets:
BookWorm \citep{papoudakis-etal-2024-bookworm}, which contains books
of various genres (primarily novels and plays) from Project Gutenberg,
and CroSS \citep{yuan-etal-2024-evaluating}, which comprises recently
published novels. Both datasets pair full-text books with 
character descriptions. Our experiments reveal that enabling the
built-in reasoning capabilities of current LLMs actually degrades
performance compared to using an empty reasoning trace. In contrast,
our approach effectively improves both the faithfulness and
informativeness of character descriptions across both datasets,
enhancing the performance of existing long-context methods. In
summary, we make the following contributions:

\vspace{-0.5ex}

\begin{itemize}[itemsep=0pt]

\item We show that the built-in reasoning mode of 
  current LLMs \emph{reduces} faithfulness for character description
  generation, with empty traces yielding more accurate outputs. 

\item We introduce QA-based guided reasoning, a modular approach that
  decouples reasoning from generation, producing a structured QA trace and then generating the final character
  description conditioned on that trace.

\item We propose a training strategy that optimizes the reasoning
  model with GRPO using gold-standard descriptions and automatically
  generated QA traces, avoiding reward design over open-ended
  generated descriptions.

\item We show that guided reasoning yields consistent improvements in
  faithfulness and informativeness across two datasets and can be 
 integrated with various long-context strategies. 
\end{itemize}


\section{Related Work}
\paragraph{Character Understanding}

Work in computational narrative understanding has focused on
characters to support story summarization~\citep{zhang2019generating},
generation~\citep{liu2020character}, analysis~\citep{10.1145/3344548},
and the simulatation of interactions among different
personas~\citep{shao-etal-2023-character}. Existing studies have
examined multiple character dimensions, including
roles~\cite{skowron2016automatic},
relationships~\cite{iyyer-etal-2016-feuding},
personality~\cite{bamman-etal-2013-learning}, and
emotions~\cite{kim-klinger-2019-frowning}. More recently, attention
has shifted to characters in longer-form narratives
(e.g.,~screenplays, books), where increased complexity (e.g., longer
plots, evolving relationships) raises modeling challenges and
motivates evaluation of long-context systems. In this setting, several
prediction-based tasks have been introduced, such as personality
prediction~\cite{sang-etal-2022-mbti, yu-etal-2023-personality} and
assigning characters to tropes in
scripts~\cite{baruah-narayanan-2025-chatter}. Previous work has also
explored richer character representations, e.g.,~by constructing
character sheets~\cite{gurung-lapata-2024-chiron} or learning
character embeddings~\citep{inoue-etal-2022-learning} from
books. Another line of research addresses character-based coreference
resolution in novels~\citep{martinelli-etal-2025-bookcoref} and
screenplays~\citep{baruah-narayanan-2023-character}.

In parallel, recent work has framed character understanding as a
text-generation problem, producing character descriptions and
analyses~\citep{brahman-etal-2021-characters-tell,
papoudakis-etal-2024-bookworm}, structured
profiles~\citep{yuan-etal-2024-evaluating}, and persona-conditioned
dialogues~\citep{chen-etal-2023-large}. We adopt a similar
setting, focusing on free-form character description generation in
book-length narratives. Unlike~\citet{yuan-etal-2024-evaluating}, we
target unstructured descriptions, as existing datasets provide
human-written text requiring no additional processing.

\paragraph{Long-context Models and Reasoning}

Language models can now process extremely long sequences, up to
millions of tokens~\cite{team2024gemini}, thanks to advances in sparse
attention~\citep{beltagy2020longformer}, long-context
post-training~\citep{gao-etal-2025-train}, scalable positional
embeddings~\citep{ding2024longropeextendingllmcontext}, and
system~\citep{kwon2023efficient} and
hardware-level~\cite{dao2022flashattention} optimizations. However,
greater context length does not reliably translate into better
long-document reasoning; recent studies report persistent failure
modes, such as position bias~\citep{liu-etal-2024-lost} and difficulty
exploiting in-context examples~\citep{li2024long}. As the training and
inference costs of long-context modeling remain a major barrier~\cite
{fu2024challenges}, previous work has also explored efficient
alternatives that handle long documents with short-context models by
incorporating retrieval, hierarchical processing, or incremental
updating~\citep{chang2023booookscore}. By operating on chunks or
selectively attending to limited context, these methods can be
competitive across a range of long-context
settings~\citep{xu2023retrieval}. In this paper, we show how guided
reasoning can be layered on top of both chunk-based approaches
(retrieval, hierarchical processing, and incremental updating) and
long-context LLMs.

Effective reasoning is also crucial for inferring implicit
information, filtering irrelevant content, and integrating evidence
across long texts. Early work elicited reasoning via
prompting~\citep{wei2022chain} or
fine-tuning~\citep{wei2021finetuned}. More recently, post-training
with reinforcement learning has become a dominant paradigm, improving
performance across a range of tasks~\citep{ouyang2022training}, such
as mathematics~\citep{uesato2022solving} and
coding~\citep{shojaee2023execution}. However, training reasoning
models for non-verifiable tasks (e.g., long-form QA and summarisation)
remains underexplored: reward design is challenging when many
outputs are acceptable and quality depends on multiple criteria.
Recent work address this by using BLUE
\citep{chang2025bleuberi}, perplexity completion~\citep{gurung2025learning}, or self-certainty~\citep{zhao2025learning} to train without external verifiers.

In contrast, we decouple reasoning from generation using a separate
reasoning model whose output trace is injected into the
generator. This allows us to optimize the reasoner directly with
reinforcement learning on its own outputs, without defining rewards
over the downstream generated text. Inspired by work showing QA pairs
serve as effective \emph{planning proxies} for
summarization~\cite{narayan-etal-2023-conditional}, we use structured
QA traces to guide generation.  However, our QA pairs function as
\emph{reasoning scaffolds} that consolidate distributed evidence about
characters, rather than as ordering constraints for generation.


\section{QA-Guided Reasoning}

We propose a modular approach that decouples reasoning from surface
realization for character description generation. Given a book (or
long narrative)~$x$ and target character~$c$, a \emph{reasoning model}
$R_{\phi}$ produces an intermediate QA-based reasoning trace~$T$, and
a \emph{generation model} $G_{\theta}$ conditions on $T$ to produce a
free-form character description~$\hat{y}$. The key idea is to
represent intermediate reasoning as \emph{typed} question-answer items
that explicitly surface salient character facts before the final
description is written.  Figure~\ref{fig:pipeline} illustrates this
architecture; we provide details below.

\subsection{QA Reasoning}
\label{sec:qa-reasoning}

Let $x$ denote a story (e.g., a book) and $c$ a target character. When
$x$ exceeds the context length of the reasoning model, we split it
into $m$ chunks $ x = [\chi_1, \ldots, \chi_m].  $ Given a chunk
$\chi_i$, the reasoning model $R_{\phi}$ outputs either \texttt{None}
(if $c$ is not mentioned in $\chi_i$) or a set of QA tuples:
\vspace{-0.5ex}
\begin{equation}
T_i \;=\; R_{\phi}(c, \chi_i) \;=\; \{(q_j, e_j, a_j, t_j)\}_{j=1}^{n_i},
\label{eq:trace-per-chunk}
\end{equation}
where $q_j$ is a question about $c$, $e_j$ is a short supporting
explanation (1--2 sentences), $a_j$ is a short answer (typically 1--4
words), and $t_j \in \mathcal{T}$ is a question type.  Following
previous work~\citep{papoudakis-etal-2024-bookworm}, we use the
question types \textit{role}, \textit{relationship},
\textit{personality}, \textit{event} and \textit{other}, which
encourage the model to generate questions with diverse topics without
focusing only on specific aspects (see
Appendix~\ref{AdditionalExperiments}).  We do not set a maximum number
of questions that the model has to generate, but instead ask it to
include all the relevant information for the character of interest (or
\texttt{None} otherwise). The final reasoning trace is the
concatenation of chunk-level traces,
$T=\textsc{Concat}(T_1,\ldots,T_m)$.

We experimented with several question-generation strategies, including
generating questions directly from the full context, generating
questions per chunk and then concatenating them, or multi-step
pipelines that first extract topics~\citep{noorbakhsh2025savaal} or
create plans~\citep{li-zhang-2024-planning} (see Appendix~\ref{AdditionalExperiments}). Since these variants
performed similarly or worse, we adopt
chunk-based QA generation for simplicity and to facilitate the training of
$R_{\phi}$ (Section~\ref{sec:training}).

\subsection{Generation Model}
\label{sec:generation-model}

The generation model $G_{\theta}$ produces a free-form character
description conditioned on both the input context and the reasoning
trace:
\vspace{-0.5ex}
\begin{equation}
\hat{y} \;=\; G_{\theta}(c, x, T).
\label{eq:gen}
\end{equation}
Operationally, we inject $T$ into the model input between special
thinking tokens (e.g.,~\texttt{<think>} $\cdots$ \texttt{</think>}),
and prompt the model to return a single-paragraph description. This
design is compatible with both reasoning-enabled LLMs and standard
instruction-tuned models that do not explicitly emit reasoning traces.

\subsection{Training}
\label{sec:training}

We finetune the $R_{\phi}$ and $G_{\theta}$ separately.  Since we do
not have gold-standard reasoning traces, we derive
\emph{silver-standard} supervision from the gold description $y$ by
extracting a reference set of QA pairs. The structured format of these
reasoning traces allows us to parse the generated QA-pairs, verify
each of them, and provide a combined reward for the entire
sequence. We employ an LLM-as-judge model to compute precision, the
percentage of generated QA-pairs that can be verified from the
gold-standard description, and recall, the percentage of gold
questions (extracted from descriptions) that the sequence of generated
QA-pairs can verify. We use the F1 score as a reward signal. Precision
and recall are defined as follows:
\vspace{-0.5ex}
\begin{equation}
R_{\text{precision}}=\frac{1}{|T|} \sum_{(q_i, a_i) \in T} \left( \textsc{verify}(q_i, a_i, P) \right)
\end{equation}
\vspace{-0.5ex}
\begin{equation}
R_{\text{recall}}=\frac{1}{|P|} \sum_{(q_i, a_i) \in P} \left( \textsc{verify}(q_i, a_i, T) \right)
\end{equation}
We optimize $R_{\phi}$ with GRPO \citep{shao2024deepseekmath}, which
samples multiple traces per prompt and performs a relative policy
update without a critic, reducing computational cost.


While we use $G_{\theta}$ in a zero-shot setting in several of our
experiments, the generator can also be \emph{fine-tuned} to better
exploit guided-QA traces. Concretely, we perform SFT on tuples
$(c, x, T, y)$ where $T$ is produced by the guided-QA reasoner (either
zero-shot or GRPO-trained), and optimize
$\log p_{\theta}(y \mid c, x, T)$ (optionally applying loss on the
injected trace tokens as well).
At inference time, the reasoning trace is always supplied by
$R_{\phi}$, and $G_{\theta}$ generates only the final description.

\subsection{Integrating Guided Reasoning with Long-Context Methods}
\label{sec:integration}

Because $R_{\phi}$ operates on chunks, our QA-guided pipeline is
model- and strategy-agnostic: it can be combined with several
long-context processing methods by producing traces $T_i$ for the same
chunks $\chi_i$ that these methods already use. Concretely, we
investigate the following settings:

\paragraph{Retrieval-based Methods}\hspace{-1ex}\cite{papoudakis-etal-2024-bookworm}
first select a character-relevant subset of the input
(e.g.,~paragraphs mentioning~$c$), yielding $x' \subseteq x$. We
simply run $R_{\phi}$ over chunks of~$x'$ and condition the generator
on the resulting trace~$T$ and retrieved context.

\vspace{-0.5ex}

\paragraph{Hierarchical Methods}\hspace{-1ex}\cite{chang2023booookscore} split the input into chunks
and generate intermediate descriptions for each chunk, which are
subsequently merged by a second-stage model. We apply guided reasoning
only at the first level: for each chunk~$\chi_i$, we generate a
trace~$T_i$ and produce an intermediate description conditioned
on~$(\chi_i, T_i)$. Merging then operates over the intermediate
descriptions without additional reasoning.  We employ the reasoning
model in a zero-shot setting and trained with GRPO, while the
generation model operates in zero-shot mode. We do not finetune the
generation model via SFT, as intermediate descriptions for individual
chunks are not available, and the model must both generate and merge
descriptions, which makes fine-tuning impractical.

\vspace{-0.5ex}

\paragraph{Incremental  Methods}\hspace{-1ex}\cite{chang2023booookscore}
process the input chunk-by-chunk, while maintaining a running global
description.  At step~$i$, the generation model updates the current
description using the new chunk and its trace,
i.e.,~$\hat{y}_i = G_{\theta}(c, \chi_i, T_i, \hat{y}_{i-1})$. This
allows newly observed evidence about the character to be integrated as
it appears in the narrative. Similarly to the hierarchical approach,
we only train the reasoning model with GRPO, while the generation
model operates in zero-shot mode.

\vspace{-0.5ex}

\paragraph{Long-context LLMs}\hspace{-1ex}can consume the full input~$x$
(within their context window) for generation. In this case, we still
produce traces chunk-wise with $R_{\phi}$ and inject their
concatenation $T$ into the long-context generator,
i.e.,~$\hat{y}=G_{\theta}(c, x, T)$.


\section{Experimental Setting}

\begin{table}[t]
\centering
\small
\begin{tabular}{lrr}
\toprule
& \multicolumn{1}{c}{\bf BookWorm} & \multicolumn{1}{c}{\bf CroSS} \\
\toprule
Books & 324 & 126 \\
Characters &  9.74 & 6.56 \\
Samples &  5,869 & 824 \\
Input Length & 97,685 & 129,113  \\
Output Length & 88.79 & 295.27 \\
\bottomrule
\end{tabular}
\caption{Dataset statistics: unique books, average
  characters per book, total examples (book–description pairs), and
  average input/output length in words.
\label{tab:datasets-stats}
\vspace{-10pt}
}
\end{table}

\paragraph{Datasets}
We use the BookWorm dataset~\cite{papoudakis-etal-2024-bookworm},
which contains books of various genres (mostly novels and plays) from
Project Gutenberg paired with characters descriptions from literature
websites.\footnote{BookWorm has two tasks (description and analysis);
we use the description partition, which contains more examples.} For
out-of-distribution evaluation, we use
CroSS~\cite{yuan-etal-2024-evaluating}, a dataset of novels published in
2022--2023 paired with character descriptions. Dataset statistics are
in Table~\ref{tab:datasets-stats}, examples in
Appendix~\ref{Examples}.

\paragraph{Model Comparisons}
We use Qwen-3-8b~\citep{yang2025qwen3} as the backbone model for all our
experiments.  We report a \textit{No Context} baseline, where the
model is prompted to generate a character description given the
character name and book title (i.e., without any story text). This
baseline estimates how much parametric knowledge the model already has
about the character, serving as a proxy for potential
contamination. We also include a \textit{Lead} baseline, truncating
the input story to the maximum length supported by the backbone model.

We evaluate our QA-guided approach on top of retrieval-augmented
methods that select character-relevant evidence and condition
generation on the retrieved context. We consider two retrieval
strategies. (a)~\textit{Coref} runs a coreference resolver to identify
chunks in which the target character is mentioned, concatenates the
selected chunks, and truncates if the resulting context exceeds the
model's input budget. We use BookNLP~\footnote{\url{https://github.com/booknlp/booknlp}}, which prior research has shown
to achieve strong off-the-shelf performance. (b)~\textit{BM25} uses
the character name as a query and retrieves chunks by BM25 score,
concatenating them until the model's context limit is reached (32k for
our experiments with Qwen-3-8b). For both retrieval approaches, we use
512-token length chunks.

We further integrate QA-guided reasoning with two chunk-based
long-context strategies, viz., \textit{Hierarchical} processing and
\textit{Incremental} updating (see Section~\ref{sec:integration}). For both methods, we use 16k-token chunks. Additional implementation
details and results are in Appendix~\ref{sec:impl-deta}
and~\ref{AdditionalExperiments}.

All methods are evaluated in three settings: without reasoning (empty
trace), with the model's built-in reasoning, and with our proposed
guided-QA trace. For guided-QA, we compare zero-shot and
GRPO-optimized versions.  For retrieval-based methods, we also train
an SFT variant, computing the loss only on the target descriptions.

\paragraph{Evaluation Metrics}

We use a set of automatic evaluation metrics to assess the quality of
generated character descriptions following previous
work~\cite{papoudakis-etal-2024-bookworm}.  

\textbf{PRISMA}~\cite{mahon-lapata-2024-modular} is an LLM-as-a-judge
metric that evaluates factual accuracy. It extracts facts from the
generated description and assesses their correctness against the gold
standard, then extracts facts from the gold standard and checks
whether the generated output supports them. These precision and recall
scores are combined into PRISMA F1. We use GPT-4o-mini for fact
extraction and Bespoke-MiniCheck-7B~\citep{tang-etal-2024-minicheck}
for fact validation, which achieves SotA performance on
LLM-AggreFact.

We also evaluate factual precision against the input story using an
entailment-based \textbf{NLI} (natural language inference) metric. For
each extracted fact, we calculate entailment scores against story
chunks, taking the maximum score across all chunks. A fact is
considered grounded if its score exceeds~0.5. We use
Bespoke-MiniCheck-7B with 1,024-token chunks for this evaluation.  We
complement NLI metrics with \textbf{QA-based
evaluation}~\citep{deutsch-etal-2021-towards} which measures whether
the generated description contains the key information needed to
answer questions derived from the reference.  We use GPT-4o-mini to
generate QA-pairs from the reference and
DeBERTaV3~\citep{he2023debertav} fine-tuned on
SQuADv2~\citep{rajpurkar-etal-2018-know} for question-answering.

We further report the \textbf{entity-mention} F1 score as a measure
entity-level coverage. Specifically, we compute precision as the
proportion of the entities mentioned in the generated description that
also appear in the gold-standard description, and recall as the
proportion of gold-standard entities recovered by the generated
output; these then combine into entity-mention F1. We also report
\textbf{Rouge-L}~\citep{lin-2004-rouge}, which measures the longest
common subsequence with the reference descriptions.

\begin{table*}[t]
\centering
\small
\begin{tabular}{ll@{\hspace{0.5em}}cc@{\hspace{1em}}ccccc}
\toprule
& \textbf{Method} & \textbf{Trace} & \textbf{Desc} & \textbf{PRISMA} & \textbf{QA} & \textbf{NLI} & \textbf{EntMent} & \textbf{Rouge-L}  \\
\midrule 
\multirow{26}{*}{\rotatebox{90}{\textbf{Qwen-3-8b}}} & No Context & --- & ZS & 11.97 & 10.31 & --- & 25.97 & 17.55 \\
& Lead-32k & --- & ZS & 15.72 & 12.35 & 61.01 & 30.60 & 17.79 \\
\cmidrule(lr){2-9}
& BM25-32k  & --- & ZS & 17.09 & 13.56 & 58.58 & 34.06 & 18.18 \\
& BM25-32k  & --- & SFT & \hspace{.4em}16.17$^{\dagger}$ & 13.26 & \hspace{.4em}60.00$^{\dagger}$ & \hspace{.4em}33.11$^{\dagger}$ & \hspace{.4em}18.99$^{\dagger}$ \\
&\quad + reasoning & {\footnotesize built-in} & ZS & 17.06 & 13.50 & 58.44 & 35.13 & \hspace{.4em}16.51$^{\dagger}$ \\
&\quad + guided-QA & ZS & ZS & 17.84 & \hspace{.4em}14.66$^{\dagger}$ & \hspace{.4em}59.96$^{\dagger}$ & \hspace{.4em}36.23$^{\dagger}$ & \hspace{.4em}17.43$^{\dagger}$\\
&\quad + guided-QA & GRPO & ZS & \hspace{.4em}18.38$^{\dagger}$ & \hspace{.4em}15.18$^{\dagger}$ & \hspace{.4em}59.90$^{\dagger}$ & \hspace{.4em}37.05$^{\dagger}$ & \hspace{.4em}17.59$^{\dagger}$ \\
&\quad + guided-QA & GRPO & SFT & \hspace{.4em}18.27$^{\dagger}$ & \hspace{.4em}15.15$^{\dagger}$ & \hspace{.4em}60.46$^{\dagger}$ & \hspace{.4em}36.60$^{\dagger}$ & \hspace{.4em}17.87$^{\dagger}$ \\
\cmidrule(lr){2-9}
& Coref-32k & --- & ZS & 19.11 & 13.96 & 52.58 & 36.12 & 18.44 \\
& Coref-32k  & --- & SFT & \hspace{.4em}17.46$^{\dagger}$ & 14.36 & 52.54 & 35.59 & \hspace{.4em}\textbf{19.26}$^{\dagger}$ \\
&\quad + reasoning & {\footnotesize built-in} & ZS & \hspace{.4em}18.59$^{\dagger}$  & 13.94 & \hspace{.4em}49.94$^{\dagger}$  & 35.88 & \hspace{.4em}16.54$^{\dagger}$ \\
&\quad + guided-QA & ZS & ZS & \hspace{.4em}18.63$^{\dagger}$ & \hspace{.4em}15.01$^{\dagger}$ & \hspace{.4em}50.60$^{\dagger}$ & 36.95 & \hspace{.4em}17.50$^{\dagger}$ \\
&\quad + guided-QA & GRPO & ZS & 19.49 & \hspace{.4em}14.86$^{\dagger}$ & \hspace{.4em}51.23$^{\dagger}$ & \hspace{.4em}37.66$^{\dagger}$ & \hspace{.4em}18.01$^{\dagger}$ \\
&\quad + guided-QA & GRPO & SFT & 19.32 & \hspace{.4em}15.10$^{\dagger}$ & 52.00 & \hspace{.4em}37.58$^{\dagger}$ & 18.42 \\
\cmidrule(lr){2-9}
& Hierarchical-16k & --- & ZS & 19.99 & 15.79 & 70.57 & 36.56 & 18.22 \\
&\quad + reasoning & {\footnotesize built-in} & ZS & 19.74 & \hspace{.4em}15.16$^{\dagger}$ & 71.15 & 35.82 & \hspace{.4em}17.33$^{\dagger}$ \\
&\quad + guided-QA & ZS & ZS & 19.82 & 16.20 & 70.74 & \hspace{.4em}37.67$^{\dagger}$ & \hspace{.4em}17.90$^{\dagger}$ \\
&\quad + guided-QA & GRPO & ZS & \hspace{.4em}18.94$^{\dagger}$ & 15.36 & \hspace{.4em}67.92$^{\dagger}$ & 36.49 & 18.19 \\
\cmidrule(lr){2-9}
& Incremental-16k & --- & ZS & 17.64 & 14.42 & 69.68 & 35.59 & 17.58 \\
&\quad + reasoning & {\footnotesize built-in} & ZS & \hspace{.4em}16.03$^{\dagger}$ & \hspace{.4em}13.36$^{\dagger}$ & \hspace{.4em}66.46$^{\dagger}$ & \hspace{.4em}33.37$^{\dagger}$ & \hspace{.4em}15.76$^{\dagger}$ \\
&\quad + guided-QA & ZS & ZS & \hspace{.4em}16.80$^{\dagger}$ & 14.56 & \hspace{.4em}68.20$^{\dagger}$ & 34.92 & \hspace{.4em}16.98$^{\dagger}$ \\
&\quad + guided-QA & GRPO & ZS & \hspace{.4em}16.59$^{\dagger}$ & \hspace{.4em}13.63$^{\dagger}$ & \hspace{.4em}66.05$^{\dagger}$ & \hspace{.4em}34.51$^{\dagger}$ & 17.42 \\
\cmidrule(lr){2-9}
& Lead-128k (w/ YaRN) & --- & ZS & 17.02 & 13.68 & 69.29 & 34.94 & 18.11 \\
&\quad + reasoning & {\footnotesize built-in} & ZS & 16.93 & 13.65 & \hspace{.4em}67.17$^{\dagger}$ & 34.20 & \hspace{.4em}16.43$^{\dagger}$ \\
&\quad + guided-QA & ZS & ZS & \hspace{.4em}19.32$^{\dagger}$ & \hspace{.4em}15.77$^{\dagger}$ & \hspace{.4em}71.55$^{\dagger}$ & \hspace{.4em}\textbf{37.80}$^{\dagger}$ & \hspace{.4em}17.77$^{\dagger}$ \\
&\quad + guided-QA & GRPO & ZS & \hspace{.4em}19.30$^{\dagger}$ & \hspace{.4em}16.22$^{\dagger}$ & \hspace{.4em}70.51$^{\dagger}$ & \hspace{.4em}37.12$^{\dagger}$ & \hspace{.4em}18.39$^{\dagger}$ \\
\midrule
\multirow[c]{2}{*}{\rotatebox[origin=c]{90}{\centering\tiny\shortstack[c]{\textbf{GPT-4.1}\\\textbf{mini}}}}
& No context & --- & ZS & 17.36 & 12.49 & --- & 29.35 & 17.41 \\
& Full context & --- & ZS & \textbf{22.05} & \textbf{16.67} & \textbf{75.53} & 35.41 & 17.71 \\
\bottomrule
\end{tabular}
\caption{Results on BookWorm.
  Trace: --- (none), built-in (model's default), ZS (zero-shot
  guided-QA), GRPO (GRPO-trained guided-QA). Desc: ZS (zero-shot) or
  SFT (supervised fine-tuning). $^{\dagger}$indicates statistically significant difference
  from the corresponding baseline without reasoning (i.e., first row
  for each method group); \textbf{bold} indicates best per metric.}
\label{bookworm_main}
\vspace{-10pt}
\end{table*}

To evaluate the \emph{quality of the QA-guided trace}, we compare
QA-pairs produced by the reasoning model against QA-pairs derived from
reference descriptions. We measure whether each generated pair is
supported by the gold QA set (precision), and how many gold QA pairs
are covered by the generated set (recall). We use GPT-4o-mini in an
LLM-as-a-judge setting for this evaluation. Our prompts and further
details are in Appendix~\ref{sec:impl-deta}.


\section{Results}

Table~\ref{bookworm_main} reports performance on BookWorm across
retrieval, hierarchical, incremental, and long-context settings, with
and without reasoning.

\paragraph{Built-in reasoning hurts faithfulness.}
Across settings, enabling the model's \emph{built-in} reasoning
consistently reduces overall quality compared to an empty trace. For
example, under Coref-32k, built-in reasoning lowers Rouge-L (18.44 $\rightarrow$
16.54) and NLI (52.58 $\rightarrow$ 49.94). A similar
pattern holds for Hierarchical-16k, where built-in reasoning decreases
QA (15.79 $\rightarrow$ 15.16) and Rouge-L (18.22 $\rightarrow$
17.33). The effect is also present in the long-context setting for Lead-128k,
with NLI dropping from 69.29 to 67.17 and Rouge-L from 18.11 to
16.43. These results suggest that default ``thinking'' traces
introduce verbosity or unsupported details, harming grounding in long
narratives (see Table~\ref{stats} in
Appendix~\ref{AdditionalExperiments} for discussion on reasoning
traces).

\paragraph{QA-guided reasoning improves grounding and coverage.}
In contrast, our guided-QA traces improve evidence-sensitive metrics,
particularly QA F1 and entity coverage. For BM25-32k, guided-QA yields
substantial gains in EntMent (34.06 $\rightarrow$ 36.23) and QA (13.56
$\rightarrow$ 14.66), while also improving NLI (58.58 $\rightarrow$
59.96), indicating better grounding in the retrieved evidence. For
Coref-32k, guided-QA improves QA (13.96 $\rightarrow$ 15.01) while maintaining comparable
EntMent. For hierarchical processing, guided-QA provides gains in EntMent (36.56
$\rightarrow$ 37.67), maintains comparable QA F1 and PRISMA, demonstrating that reasoning helps even when the
model already aggregates chunk-level summaries. In the long-context
setting, guided-QA with GRPO-trained traces improves over the baseline
across all metrics except R-L (PRISMA: 17.02 $\rightarrow$ 19.30, QA: 13.68
$\rightarrow$ 16.22). Overall, guided-QA is most beneficial when the
context selection step surfaces relevant evidence, but the generator
needs help integrating it into a coherent, grounded description.

\begin{table*}[t]
\centering
\small
\begin{tabular}{llcc@{\hspace{0.8em}}ccccc}
\toprule
\textbf{Model} & \textbf{Method} & \textbf{Trace} & \textbf{Desc} &  \textbf{PRISMA} & \textbf{QA} & \textbf{NLI} & \textbf{EntMent} & \textbf{Rouge-L}  \\
\midrule 
\multirow{9}{*}{\shortstack[l]{Qwen-3-8b}} 
& No Context & --- & ZS & \hspace{1.2ex}7.69 & \hspace{1.2ex}2.92 & --- & 19.83 & 16.40 \\
& Coref-32k & --- & ZS & 23.13 & 13.03 & 50.01 & 35.90 & 17.44 \\
& \quad + reasoning & {\footnotesize built-in} & ZS & \hspace{.9ex}22.26$^{\dagger}$ & 12.52 & \hspace{.9ex}46.84$^{\dagger}$ & \hspace{.9ex}33.50$^{\dagger}$ & \hspace{.9ex}16.02$^{\dagger}$ \\
& \quad + guided-QA & ZS & ZS & \hspace{.9ex}21.88$^{\dagger}$ & 13.11 & 49.63 & 35.46 & \hspace{.9ex}17.24$^{\dagger}$ \\
& \quad + guided-QA & GRPO & ZS & \hspace{.9ex}22.63$^{\dagger}$ & \hspace{.9ex}13.08$^{\dagger}$ & 49.70 & \hspace{.9ex}35.38$^{\dagger}$ & \hspace{.9ex}17.33$^{\dagger}$  \\
\cmidrule(lr){2-9}
& Lead-128k (w/ YaRN) & --- & ZS & 22.22 & 13.70 & 66.39 & 35.48 & 17.21 \\
& \quad + reasoning & {\footnotesize built-in} & ZS & \hspace{.9ex}18.08$^{\dagger}$ & \hspace{.9ex}10.97$^{\dagger}$ & 66.98 & \hspace{.9ex}31.18$^{\dagger}$ & \hspace{.9ex}15.75$^{\dagger}$ \\
& \quad + guided-QA & ZS & ZS & \hspace{.9ex}23.72$^{\dagger}$ & \hspace{.9ex}14.62$^{\dagger}$ & \hspace{.9ex}68.57$^{\dagger}$ & \hspace{.9ex}\textbf{36.84}$^{\dagger}$ & \hspace{.9ex}17.47$^{\dagger}$ \\
& \quad + guided-QA & GRPO & ZS & \hspace{.9ex}23.78$^{\dagger}$ & \hspace{.9ex}14.42$^{\dagger}$ & \hspace{.9ex}68.09$^{\dagger}$ & \hspace{.9ex}36.18$^{\dagger}$ & \hspace{.9ex}\textbf{17.49}$^{\dagger}$ \\ 
\midrule
\multirow{2}{*}{\shortstack[l]{GPT-4.1-mini}} 
& No context & --- & ZS & 10.01 & \hspace{1.2ex}3.98 & --- & 21.46 & 16.10 \\
& Full context & --- & ZS & \textbf{28.49} & \textbf{16.08} & \textbf{72.67} & 28.49 & 16.80 \\
\bottomrule
\end{tabular}
\caption{Results on CroSS. Trace: --- (none), built-in (model's
  default), ZS (zero-shot guided-QA), GRPO (GRPO-trained
  guided-QA). Desc: ZS (zero-shot) or SFT (supervised
  fine-tuning). $^{\dagger}$indicates statistically significant difference from the
  corresponding baseline without reasoning (i.e., first row for each
  method group); \textbf{bold} indicates best per metric.}
\label{cross_main}
\vspace{-10pt}
\end{table*}

\vspace{-1ex}
\paragraph{Trace quality and generator training affect different metrics.}
Comparing zero-shot and GRPO-trained traces reveals that better traces translate into better descriptions, though gains are metric-dependent. Under BM25-32k, GRPO traces improve PRISMA (17.09 $\rightarrow$ 18.38) while preserving high EntMent, whereas SFT on descriptions primarily benefits surface-overlap metrics (e.g., Rouge-L: 18.18 $\rightarrow$ 18.99) with weaker effects on factual scores. A similar pattern emerges for Coref-32k, where SFT achieves the highest Rouge-L (19.26) but GRPO-trained traces yield the best QA F1 (14.72). This suggests that trace-level optimization and generation-level training target complementary aspects of output quality.

\paragraph{Benefits in incremental settings are limited.}
Incremental processing presents a challenging setting where guided-QA
shows limited benefit. Both built-in and guided reasoning degrade
PRISMA compared to the baseline (17.64 $\rightarrow$ 16.03 and 16.80,
respectively). We hypothesize that the sequential updating mechanism
struggles to incorporate reasoning traces effectively, as each update
step must reconcile new QA pairs with a partial description. This
suggests that our approach is best suited for settings where evidence
can be aggregated before generation rather than integrated
incrementally.

\paragraph{Entity coverage improves even against proprietary models.}
For reference, we include GPT-4.1-mini with full context access, which
achieves the highest PRISMA (22.05), QA F1 (16.67) and NLI (75.53). However,
guided-QA with the smaller Qwen-3-8b model narrows this gap,
achieving the best EntMent score overall (37.80 vs. 35.41). This
indicates that structured reasoning traces improve entity coverage
even compared to larger proprietary models with longer context
windows.

\begin{table}[t]
    \small
    \centering
    {\setlength{\tabcolsep}{3pt}
\resizebox{\columnwidth}{!}{%
\begin{tabular}{@{}l@{~}c@{~}c@{~}c@{~}c@{~}|@{~}c@{~}c@{}}
\toprule
& \multicolumn{4}{c}{\textbf{QA reasoning}} &  \multicolumn{2}{c}{\textbf{Description}} \\
{\textbf{Method}} & \# QA &  {Prec} & {Rec} & \multicolumn{1}{c@{~~}}{F1}  & \multicolumn{1}{c}{PRISMA} & QA\\
\midrule 
No reasoning & --- & --- & --- & --- & 15.20 & 14.42 \\
No chunking  & 6.50 & \textbf{16.10} & 16.93 & 16.50 & 14.02 & 14.03 \\
guided-QA  & 8.60 & 15.07 & 17.68 & 16.27 & 14.45 & 14.45\\
guided-QA (SFT) & 6.60 & 14.84 & 15.98 & 15.38 & 14.11 & 14.08 \\
guided-QA (GRPO) & 11.07 & 15.25 & \textbf{19.10} & \textbf{16.96} & \textbf{15.25} & \textbf{14.79} \\
Oracle & 7.51 & --- & --- & --- & 44.81 & 40.61 \\
\bottomrule
\end{tabular}
}
 \caption{Comparison of question generation methods for the reasoning
   model on the BookWorm dataset (validation set) with coreference-based retrieval. The generation model is zero-shot in all experiments.
\label{qa_prof_corr}}}
\vspace{-15pt}
\end{table}

\paragraph{Faithful QA traces improve descriptions across QA strategies.}
Table~\ref{qa_prof_corr} compares alternative question-generation
strategies for the reasoning model and relates trace quality to
downstream description performance (the generator is kept zero-shot in
all settings). We also report an \textit{Oracle} upper bound, where QA
pairs are extracted from the gold descriptions and used directly as
traces, highlighting the remaining headroom when perfect intermediate
signals are available.

Overall, optimizing the reasoner improves both trace quality and
description quality. In particular, GRPO yields the best trace
recall and the highest trace F1 (16.96), outperforming both
zero-shot guided reasoning and SFT on traces. This improvement
translates downstream: conditioning on GRPO traces produces the best QA F1 (14.79) and PRISMA score (15.25), though the PRISMA gains over no reasoning are only marginal (15.20 $\rightarrow$ 15.25). By contrast, increasing
the number of QA pairs alone is not sufficient: guided-QA
generates more questions than the no chunking approach but has lower precision and total F1 score.
These results suggest that downstream gains are driven by
\emph{faithful} and \emph{informative} traces rather than by trace
length (see question generation experiments in Appendix~\ref{AdditionalExperiments}). Finally, the oracle traces dramatically outperform all
automatic traces, indicating substantial room for improvement in trace
generation and verification.

\vspace{-1ex}
\paragraph{Guided reasoning improves transfer to CroSS.}
Table~\ref{cross_main} evaluates whether models tuned on BookWorm
transfer to CroSS~\citep{yuan-etal-2024-evaluating}. Using Qwen-3-8b,
character-focused context strategies yield substantial gains over no
context (e.g., Coref-32k PRISMA: 7.69$\rightarrow$23.13). Consistent
with BookWorm, built-in reasoning degrades grounding metrics
(Coref-32k NLI: 50.01 $\rightarrow$ 46.84; Lead-128k EntMent: 35.48
$\rightarrow$ 31.18), while guided-QA improves or keeps comparable performance across
settings. For Lead-128k, guided-QA with GRPO traces achieves the best
PRISMA (23.78) and QA F1 (14.42) among Qwen-3-8b configurations,
demonstrating that our approach benefits long-context processing on
unseen data. GPT-4.1-mini with full context provides a strong upper
bound (PRISMA 28.49, NLI 72.67), though Qwen-3-8b with guided
reasoning achieves competitive or superior entity coverage (EntMent:
36.84 vs. 28.49), confirming that structured traces improve grounding
even against larger models.


\section{Conclusion}
We address the challenge of generating accurate character descriptions
from book-length narratives by proposing a modular framework that
decouples reasoning from generation. Our experiments reveal that
enabling the built-in reasoning mode of current LLMs often degrades
performance on character description generation, with empty reasoning
traces yielding more faithful outputs across multiple long-context
settings.  To address this, we introduce QA-guided reasoning, a
two-stage approach consisting of (1) a reasoning model that generates
structured question-answer traces capturing salient character
information, and (2) a generation model that conditions on these
traces to produce final descriptions.

A key advantage of our approach is its training strategy: the
reasoning model is optimized directly with Group Relative Policy
Optimization on gold-standard character descriptions and automatically
derived QA traces, eliminating the need to define rewards over the
final generated text, which is a particularly challenging problem for
open-ended generation tasks. The generation model can be trained with
supervised fine-tuning by injecting reasoning traces between thinking
tokens.  Experiments on two datasets (BookWorm and CroSS) demonstrate
that QA-guided reasoning improves faithfulness and informativeness
across multiple evaluation metrics. Our analysis shows that trace
quality directly impacts downstream description quality, with
GRPO-optimized traces yielding the best performance. The approach also
transfers effectively to out-of-distribution data, suggesting that the
learned reasoning patterns generalize beyond the training
distribution.
 
Future work should explore alternative reasoning structures beyond QA
pairs, investigate other reinforcement learning algorithms and reward
formulations, and develop character-specific evaluation metrics that
better capture narrative understanding.


\section*{Limitations}
Our evaluation relies on automatic metrics following established
practices in prior work. We employ question-answering and fact-based
metrics (PRISMA) to assess descriptions against gold standards,
entailment-based metrics (NLI) to measure grounding in the input
story, and standard surface-level metrics such as entity-mention F1
and Rouge-L. While these metrics provide useful signals, they have
important limitations for evaluating long-context generation. Human
evaluation, though more reliable, remains prohibitively expensive and
difficult to scale to the thousands of examples required for robust
assessment. Automatic metrics, conversely, struggle to capture nuanced
aspects of character understanding, such as narrative coherence,
implicit trait inference, and the integration of evidence across
distant spans. Future work should develop character-specific
evaluation frameworks that better capture narrative understanding and
can be applied at scale. 

Our training approach uses GRPO to optimize the reasoning model. While
GRPO is computationally efficient and performs well in our
experiments, other reinforcement learning algorithms (e.g., PPO, DPO)
or alternative reward formulations may yield further improvements. We
leave exploration of these alternatives to future work. 

Finally, our approach requires silver-standard QA traces derived from
gold descriptions during training. In settings where high-quality
character descriptions are unavailable, alternative supervision
strategies (e.g., weak supervision from plot summaries or character
wikis) may be necessary. Investigating such strategies would broaden
the applicability of our framework.


\section*{Acknowledgments}
This work was supported in part by the UKRI Centre for Doctoral Training in Natural Language
Processing, funded by the UKRI (grant EP/S022481/1) and the University of Edinburgh, School of Informatics and School of Philosophy, Psychology \& Language Sciences. Computing resources were provided by the Edinburgh International Data Facility (EIDF) and the Data-Driven Innovation Programme at the University of Edinburgh. Access to EIDF was facilitated through the University of Edinburgh’s Generative AI Laboratory GAIL Fellow scheme.

\newpage
\clearpage

\bibliography{anthology, custom}

\appendix

\newpage
\clearpage
\appendix

\section{Implementation Details}
\label{sec:impl-deta}

\paragraph{Training and Inference}
We use the OpenRLHF~\citep{hu2024openrlhf} library for GRPO training with the hyperparameters listed in Table~\ref{grpo_hyperparam}. We also employ the Qwen-3-8b model to provide LLM-as-a-judge rewards during GRPO-training. For supervised finetuning, we use LoRA~\citep{hu2022lora} from the unsloth~\citep{unsloth} library. We report the supervised finetuning hyperparameters in Table~\ref{sft_hyperparam}. For inference, we use the vLLM engine~\citep{kwon2023efficient} with sample decoding and temperature $0.4$. We evaluate the statistical significance of the results using two-sided paired approximate randomization test (10,000 permutations and $\alpha = 0.05$)~\cite{noreen1989computer}, based on the mean value across 4 inference seeds. We used four H200 Nvidia GPUs for GPRO training and a single H100 or H200 GPU for all the inference experiments and SFT training.

\begin{table}[h]
\centering
\small
\begin{tabular}{l l}
\hline
\textbf{Hyperparameter} & \textbf{Value} \\
\hline
actor learning rate & $5 \times 10^{-7}$ \\
kL coefficient & $0.01$ \\
train batch size & $64$ \\
samples per prompt & $8$ \\
prompt max length & $17,684$ \\
generation max length & $2,048$ \\
\hline
\end{tabular}
\caption{Training hyperparameters for Group Relative Policy Optimization (GRPO) training.}
\label{grpo_hyperparam}
\end{table}

\begin{table}[h!]
\centering
\small
\begin{tabular}{l l}
\hline
\textbf{Hyperparameter} & \textbf{Value} \\
\hline
learning rate & $10^{-6}$ \\
max input length & $32,768$ \\
batch size & $1$ \\
gradient accumulation steps & $8$ \\
alpha & $256$ \\
rank & $128$ \\
lora dropout & $0.1$ \\
\hline
\end{tabular}
\caption{Training hyperparameters for supervised finetuning (SFT) training with LoRA.}
\label{sft_hyperparam}
\end{table}

\begin{table}[t]
\centering
 \footnotesize
\begin{tabular}{@{}p{7cm}@{}}
\toprule
Context: \texttt{\{context\}} \\ \\

Describe character \texttt{\{character\}} from book \texttt{\{book\}}
based on the given context. \\ 
Return your output as a single paragraph (close to \texttt{\{length\}}
words) including the important information. \\ 
\bottomrule
\end{tabular}
\caption{Character description generation prompt template. We replace the variable length with the average output length of the corresponding dataset.}
\label{description_generation}
\end{table}

\begin{table}[h!]
\centering
 \small
\begin{tabular}{p{7cm}}
\toprule
Context: \texttt{\{context\}}
\\ \\
Your task is to generate questions answer pairs about character: \texttt{\{character\}} from book: \texttt{\{book\}} given the above chunk of the book. You should focus on understanding aspects of the character (e.g., role, relationships, personality, events) that are mentioned in the context. Each qa-pair should be labelled as role, relationship, personality, event or other. We provide the definitions for these below. \\ \\

Definitions: \\
Role: defines what part the character plays in the story, narrator, major/minor character. \\ 
Relationship: connections the character has with others, such as friendships or family ties. \\
Personality: character’s behavior, traits, and attributes. \\
Event: actions and decisions the character is involved in throughout the story. \\
Other: any other fact that doesn’t belong to the above categories. \\
\\
Output format: \\
Q1: <question> E1: <explanation> A1: <answer> T1: <type> \\
Q2: <question> E2: <explanation> A2: <answer> T2: <type> \\
... 

\\ \\
Generate an explanation, 1-2 sentences that fully justify your answer, do not simply repeat the answer. Type of qa has to be one of Role, Relationship, Personality, Event or Other. The answer should be short 1-4 words. Generate QA-pairs only related to character: \texttt{\{character\}}. Generate QA-pairs only for information mentioned in the provided context. Do not include unanswered questions. \\ \\

The questions has to mention the name of the character: \texttt{\{character\}}. Do not generate repetitive QA-pairs with same answer. If the character is not mentioned, simply return None. \\
\bottomrule
\end{tabular}
\caption{QA generation prompt template. We adopt the definitions for the different categories from~\citet{papoudakis-etal-2024-bookworm}}
\label{qa_generation}
\end{table}

\paragraph{Prompts}

We provide the prompts used for our experiments in
Tables~\ref{description_generation} and \ref{qa_generation}.

\paragraph{Evaluation}
We use the VeriScore~\citep{song-etal-2024-veriscore} codebase to extract facts (prompts are adjusted based on~\citet{papoudakis-etal-2024-bookworm}) for both PRISMA and NLI metrics. Fact verification is performed using the MiniCheck~\citep{tang-etal-2024-minicheck} model in both metrics. 
We use rouge-score\footnote{\url{https://github.com/google-research/google-research/tree/master/rouge}} implementation for Rouge-L metric and NLTK\footnote{\url{https://www.nltk.org/}} toolkit for named-entity extraction in entity-mention F1 calculation.

\section{Additional Experiments and Statistics}
\label{AdditionalExperiments}

\paragraph{QA reasoning ablation}
We run an ablation study to demonstrate the effect of the different
components in the reasoning trace. We use the Qwen-3-8B model as a
backbone with a coreference-based retrieval method to extract context
and employ the guided reasoning method in zero-shot mode. 

\begin{table}[t]
\centering
\resizebox{\columnwidth}{!}{%
\begin{tabular}{@{}l@{\hspace{0.6em}}c@{~}c@{~}c@{~}c@{\hspace{0.8em}}|c@{~}c@{~}c@{}}
\toprule
& \multicolumn{4}{c}{\textbf{QA Reasoning}} & \multicolumn{3}{c}{\textbf{Description}} \\
{\textbf{Method}} & \#QA & P & R & \multicolumn{1}{c@{\hspace{0.8em}}}{F1} & \multicolumn{1}{c@{~}}{PRI} & QA & Ent \\
\midrule 
guided-QA & 8.60 & \textbf{15.07} & \textbf{17.68} & \textbf{16.27} & \textbf{14.45} &14.45 & \textbf{30.41} \\
\quad w/o expl. & 12.05 & 13.89 & 16.93 & 15.26 & 14.34 & \textbf{15.01} & 29.94 \\
\quad w/o types & 11.44 & 12.46 & 15.87 & 13.95 & 13.41 & 14.34 & 30.37 \\
\quad w/o expl., types & 15.46 & 13.90 & 12.16 & 13.61 & 13.71 & 14.92 & 30.08 \\
\quad w/o expl., ans., types & 13.44 & --- & --- & --- & 14.02 & 14.22 & 29.62 \\
\bottomrule
\end{tabular}
}
\caption{Ablation of QA reasoning components using Qwen-3-8b with
  coreference retrieval on BookWorm validation set (zero-shot). P/R/F1 = reasoning
  precision/recall/F1; PRI = PRISMA F1; Ent = EntMent F1.} 
\label{ablation}
\end{table}

Table~\ref{ablation} examines the contribution of each component in
the QA reasoning trace: explanations, question types, and answers.
The full Guided QA method achieves the best PRISMA (14.45) and EntMent
(30.41), indicating that all components contribute to faithful and
entity-rich descriptions. Removing explanations (w/o expl.) increases
the number of generated QA pairs (8.60 $\rightarrow$ 12.05) and
improves QA F1 (14.45 $\rightarrow$ 15.01), but reduces precision  (15.07 $\rightarrow$ 13.89) and recall (17.68  $\rightarrow$ 16.93),
suggesting that explanations help maintain informative and precise reasoning traces.

Removing question types (w/o types) causes a drop in
reasoning F1 (16.27 $\rightarrow$ 13.95) and PRISMA (14.45
$\rightarrow$ 13.41), indicating that typed questions encourage
topical diversity and more informative traces. Removing both
explanations and types (w/o expl., types) generates the most QA pairs
(15.46) but with lower precision (13.90) and recall (12.16), yielding mixed downstream
results.

Finally, retaining only questions (w/o expl., ans., types) prevents
reasoning evaluation (no answers to verify) and degrades PRISMA
(14.02), confirming that answers are essential for grounding the
generation model. Overall, each component serves a distinct role:
types encourage diversity, explanations improve precision, and answers
provide the factual content that guides description generation.

\paragraph{Reasoning Statistics}
Table~\ref{stats} reports summary statistics for intermediate
reasoning traces and the final character descriptions when
conditioning on the retrieved context in BookWorm. Overall, the
\emph{description} outputs remain similar in length across settings
(approximately 109--118 tokens) and exhibit comparable lexical
diversity (about 71--73\% unique unigrams) for the no-reasoning
baseline and both guided-QA variants. This suggests that performance
differences in our main experiments are unlikely to be driven by
longer or more lexically diverse descriptions, but instead by the
content and structure of the intermediate reasoning signal. 

\begin{table}[t]
\centering
\small
\scalebox{0.98}{
\begin{tabular}{lccc|rc}
\toprule
& \multicolumn{3}{c}{\textbf{Reasoning}} & \multicolumn{2}{c}{\textbf{Description}} \\
\textbf{Method} & \#QA & Tok &  \multicolumn{1}{c}{Uni} & Tok & Uni \\
\midrule 
None & --- & --- & --- & 111 & 71 \\
built-in & --- & 270 & 60 & 109 & 78 \\
guided-QA (ZS) & 8 & 391 & 36 & 118 & 73 \\
guided-QA (GRPO) & 11 & 535 & 31 & 111 & 71 \\
\bottomrule
\end{tabular}}
\caption{Statistics for  generated traces and their corresponding
  descriptions for different reasoning methods using retrieved
  context on BookWorm validation set. \#QA = QA pairs; Tok = avg.\ tokens; Uni =
  unique unigrams (\%).} 
 \label{stats}
\end{table}

In contrast, the \emph{reasoning} traces differ
substantially. Built-in reasoning produces moderately long traces (270
tokens) with higher unigram diversity than guided-QA, whereas
guided-QA traces are much longer (391--535 tokens) but show markedly
lower unigram diversity (31--36\%), which is largely attributable to
the structured QA format (e.g., repeated question templates and answer
markers). Comparing the two guided-QA variants, GRPO yields longer
traces (11 QA pairs; 535 tokens) than the zero-shot reasoner
(8 QA pairs; 391 tokens), while maintaining similar diversity,
suggesting that optimization encourages the model to select more,
higher-yield QA pairs but without increasing verbosity. Finally,
built-in reasoning leads to the highest description unigram diversity
(78\%) despite lower faithfulness in our main results, consistent with
the interpretation that unstructured deliberation may introduce
additional (and potentially unsupported) details rather than improving
grounding.

\paragraph{Experiments with Gemma-3}
We further evaluate our approach using the Gemma-3-12b-it~\citep{team2025gemma} model for coreference-based retrieval and Lead-128k methods on the BookWorm dataset. We report a reasoning experiment in which the model is prompted to explicitly reason, generating a chain-of-thought inside ~\texttt{<think>} $\cdots$ \texttt{</think>} tokens before producing the character description. Although the Gemma-3 model is not trained to explicitly reason, we found that it consistently follows the reasoning format in all the experiments. The results in Table~\ref{gemma_results} demonstrate that explicit reasoning harms performance, for example, both PRISMA (11.76 $\rightarrow$ 11.16) and QA-F1 (11.78 $\rightarrow$ 10.67) degrade for Lead-128k. In contrast, zero-shot QA-guided reasoning in the coreference method improves performance over the corresponding approaches in all metrics except PRISMA, achieving also the highest QA-F1 score (13.60). Similarly, QA-guided reasoning for Lead-128k improves over the corresponding baselines across all metrics, achieving the highest EntMent (29.34) and R-L (16.62). Overall, these results indicate that findings from the main experiments also apply to a model from a different family, which is not trained for explicit reasoning, and that the proposed QA-guided method can also be effective under this setting.

\begin{table}[t]
\centering
\small
\resizebox{\columnwidth}{!}{%

\begin{tabular}{lcccc}
\toprule
{\textbf{Method}} & PRISMA & QA & EntMent & R-L\\
\midrule 
Coref-32k & \textbf{13.65} & 12.49 & 27.93 & 16.20 \\
\quad + reasoning & 11.90 & 11.11 & 25.60 & 15.49\\
\quad + guided-QA & 13.10 & \textbf{13.60} & 28.82 & 16.61 \\
Lead-128k & 11.76 & 11.78 & 26.39 & 15.99 \\
\quad + reasoning &11.16 & 10.67 & 25.66 & 15.16 \\
\quad + guided-QA & 13.12 & 13.32 & \textbf{29.34} & \textbf{16.62} \\
\bottomrule
\end{tabular}}
\caption{Results on BookWorm dataset using Gemma-12b-it model. Zero-shot experiments without any reasoning trace, explicitly prompting the model to reason and QA-guided reasoning for coref-based retrieval and Lead-128k methods.}
\label{gemma_results}
\end{table}

\paragraph{QA generation methods}
We compare different question generation methods, evaluating the corresponding reasoning traces. Precisely, we tested \textbf{Savaal}~\citep{noorbakhsh2025savaal}, a multi-step topic-driven approach for QA generation in long documents. This approach first extracts topics from each chunk of the input and then uses each topic to retrieve relevant passages. The retrieved passages and the corresponding topic are finally used to generate QA-pairs. 

Next, we evaluate a plan-based QA generation method based on \textbf{PFQS} (planning first, question second)~\citep{li-zhang-2024-planning}. We prompt a model to extract sentences from the given input that serve as candidate evidence for answers. The extracted sentences are provided together with the input context to generate QA pairs. We evaluate both of these approaches using the same output format, adding an explanation and type to each QA-pair (see Table~\ref{qa_generation}). We also split the input context into chunks of~16k tokens.

Table~\ref{qa_methods} shows the evaluation of the different methods and their average number of questions. The guided-QA method yields the highest precision, recall, and F1 (16.27) despite having the lowest number of QA pairs. Savaal method obtains a moderate F1 score but performs worse than guided-QA, while plan-based generation has competitive precision, but the recall is significantly lower (11.71). Overall, these results demonstrate that simple chunk-based question generation can outperform topic- and plan-based methods, yielding more precise and informative QA-pairs even with fewer generated questions.

\begin{table}[t]
\centering
\small
\begin{tabular}{lcccc}
\toprule
{\textbf{Method}} & \# QA &  {Precision} & {Recall} & {F1}\\
\midrule 
guided-QA & 8.60 & \textbf{17.68} & \textbf{15.07} & \textbf{16.27} \\
Savaal & 12.01 & 15.23 & 13.92 & 14.54 \\
Plan-based QA & 11.90 & 17.34 & 11.71 & 13.98 \\
\bottomrule
\end{tabular}
\caption{Experiments with different QA generation methods using coref-based retrieval on BookWorm dataset. We report the average number of QA-pairs along with precision, recall and F1 score for reasoning evaluation.}
\label{qa_methods}
\end{table}

\section{Examples}
\label{Examples}

We show examples of generated outputs and reasoning traces in Tables~\ref{cross_example_1},~\ref{reasoning_example_1},~\ref{tab:eval_examples_2} and~\ref{reasoning_example_2}.
\clearpage

\begin{table*}[ht]
\centering
\begin{tabular}{@{}p{0.7\textwidth} p{0.1\textwidth} p{0.1\textwidth}@{}}
\toprule
\small{}{\textbf{Character:}  Demon Copperhead \quad  \textbf{Book:} Demon Copperhead}\\
\midrule
\multicolumn{3}{@{}p{\textwidth}@{}}{\small{\textbf{Lead-128k:} Demon Copperhead is a complex and resilient character shaped by a life of hardship, neglect, and survival. Born to a drug-addicted mother, he was abandoned in a trailer park in Lee County, Virginia, and nearly died at birth due to his mother's drug use, which led to his being born in the caul—a rare occurrence that his mother claimed was a sign of her own superstitions. He was taken in by the Peggots, who were not his real family but became his de facto parents, giving him a sense of belonging and a nickname, "Demon," that stuck with him as he grew. His early life was marked by instability, including being sent to live with the brutal Crickson family on a farm, where he endured harsh treatment and emotional abuse. He also experienced foster care with Miss Barks, who became a maternal figure but eventually left for a better-paying job as a teacher. Demon's life took a tragic turn when his mother died in an overdose, leaving him in foster care again and further alienated. He later ran away to Nashville, where he encountered a series of misfortunes, including theft, abuse, and being caught in a pawnshop with a hooker. Eventually, he found refuge with his grandmother Betsy Woodall, who had long been estranged from him but now took him in after learning of his mother's death. Demon is marked by his red hair and blue eyes, traits he inherited from his father—a Melungeon man he never knew but who left an indelible mark on his identity. He is also defined by his survival instincts, having learned to navigate the harsh realities of poverty, neglect, and abuse.
} }\\
\midrule
\multicolumn{3}{@{}p{\textwidth}@{}}{\small{\textbf{Lead-128k + guided-QA (GRPO):}
Demon Copperhead, the protagonist of the novel, is a resilient and determined young man who navigates a life marked by hardship, neglect, and a complex web of relationships. Born to an addicted mother, he faces a difficult start in life, including a traumatic birth and early experiences with foster care. His early years are shaped by the Peggots, who become a significant part of his life, though their relationship is strained and complicated. He develops a close friendship with Maggot, another foster child, and later forms a bond with Fast Forward, who becomes a mentor and friend. Demon's nickname "Demon" is given to him by others, reflecting his troubled past and the stigma he carries. His relationship with Stoner, his stepfather, is fraught with tension and abuse, leading to a confrontation that results in Demon's injury and eventual placement in foster care. He is placed with Mr. Crickson, a harsh farmer who runs a slave-boy farm, but eventually finds a more supportive home with Coach Winfield and his family in Jonesville. Demon's personality is marked by his determination to prove his worth despite his circumstances, as well as his artistic talent for drawing superheroes and comic strips. He has a complex relationship with his biological mother, who dies from an overdose, and his biological father, whose death he never fully understands. His journey includes running away from home, stealing a tree for Christmas, and eventually finding love and support with Dori, who becomes his girlfriend and later his fiancée. Demon's story is one of resilience, as he struggles with addiction but ultimately seeks redemption through his art and relationships. His character is defined by his search for belonging and self-worth in a world that often overlooks him.
}} \\
\midrule
\multicolumn{3}{@{}p{\textwidth}@{}}{\small{\textbf{Goldstandard Description:}  Demon Copperhead is the eponymous, first-person narrator and protagonist of the novel. Because the novel's narrative voice deals with both the events of Demon's life and his reflection on those events, the novel is structured like a memoir. It is also a bildungsroman in which Demon endures several life-altering conflicts and eventually finds peace with himself. The world has been cruel to Demon, and his early life demonstrates The Failure of Society to Protect Its Children. He was born without a father, and his mother struggled with addiction throughout her short life. He navigates several different forms of false families, such as life with his abusive stepfather, placements in two abusive foster homes, and rejection from his father's mother. The lack of love in Demon's childhood and early adolescence makes him radically self-reliant, but it also makes him distrustful of the people who do provide him with security and care. Demon is independent because he must be to survive, but he looks for people to believe in—often misguidedly, as with Fast Forward and Dori. Similarly, he has an attachment to Lee County because it is a space he at least understands, even if it keeps bringing him down. Demon eventually learns that he must focus on himself to receive real love from other people.  [truncated]

\vspace{0.5em}

\begin{tabular}{@{}p{0.6\textwidth} >{\centering\arraybackslash}p{0.16\textwidth} >{\centering\arraybackslash}p{0.19\textwidth}@{}}

{\footnotesize\textbf{Fact}} & \textbf{\footnotesize{Lead-128k}} & \textbf{\footnotesize {Lead-128k guided-QA}} \\

Demon Copperhead is the eponymous narrator of the novel. & \textcolor{wrong}{\textbf{False}} & \textcolor{correct}{\textbf{True}} \\ 
Demon Copperhead is the first-person narrator of the novel. & \textcolor{wrong}{\textbf{False}} & \textcolor{wrong}{\textbf{False}} \\ 
Demon Copperhead is the protagonist of the novel. & \textcolor{wrong}{\textbf{False}} & \textcolor{correct}{\textbf{True}} \\ 
The narrative voice of the novel "Demon Copperhead" deals with the events of Demon's life. & \textcolor{wrong}{\textbf{False}} & \textcolor{correct}{\textbf{True}} \\ 
The narrative voice of the novel "Demon Copperhead" includes Demon's reflection on those events. & \textcolor{wrong}{\textbf{False}} & \textcolor{wrong}{\textbf{False}} \\ 
The novel "Demon Copperhead" is structured like a memoir. & \textcolor{wrong}{\textbf{False}} & \textcolor{wrong}{\textbf{False}} \\ 
"Demon Copperhead" is a bildungsroman. & \textcolor{wrong}{\textbf{False}} & \textcolor{correct}{\textbf{True}} \\ 
In "Demon Copperhead," Demon endures several life-altering conflicts. & \textcolor{correct}{\textbf{True}} & \textcolor{correct}{\textbf{True}} \\ 
In "Demon Copperhead," Demon eventually finds peace with himself. & \textcolor{wrong}{\textbf{False}} & \textcolor{wrong}{\textbf{False}} \\ 
The world has been cruel to Demon Copperhead. & \textcolor{correct}{\textbf{True}} & \textcolor{correct}{\textbf{True}} \\ 
Demon's early life demonstrates The Failure of Society to Protect Its Children. & \textcolor{correct}{\textbf{True}} & \textcolor{correct}{\textbf{True}} \\ 
Demon was born without a father. & \textcolor{correct}{\textbf{True}} & \textcolor{wrong}{\textbf{False}} \\ 
Demon's mother struggled with addiction throughout her short life. & \textcolor{wrong}{\textbf{False}} & \textcolor{correct}{\textbf{True}} \\ 
Demon navigates several different forms of false families. & \textcolor{correct}{\textbf{True}} & \textcolor{correct}{\textbf{True}} \\ 
Demon lived with an abusive stepfather. & \textcolor{wrong}{\textbf{False}} & \textcolor{correct}{\textbf{True}} \\ 
Demon was placed in two abusive foster homes. & \textcolor{correct}{\textbf{True}} & \textcolor{correct}{\textbf{True}} \\ 
Demon experienced rejection from his father's mother. & \textcolor{wrong}{\textbf{False}} &  \textcolor{wrong}{\textbf{False}} \\ 
The lack of love in Demon’s childhood made him radically self-reliant. & \textcolor{correct}{\textbf{True}} & \textcolor{correct}{\textbf{True}} \\ 
{[Facts truncated]} & & \\
\end{tabular}
}} \\
\bottomrule
\end{tabular}

\caption{Example of generated descriptions in CroSS dataset with QA-guided Lead-128k and the corresponding baseline. We demonstrate facts evaluated based on the PRISMA-recall metric.}
\label{cross_example_1}
\end{table*}

\begin{table*}[ht]
\centering
\begin{tabular}{@{}p{0.7\textwidth} p{0.1\textwidth} p{0.1\textwidth}@{}}
\toprule
\small{}{\textbf{Character:}  Demon Copperhead \quad  \textbf{Book:} Demon Copperhead}\\
\midrule
\multicolumn{3}{@{}p{\textwidth}@{}}{
\small
\textbf{Reasoning for 1st Chunk:}
\vspace{0.5em}

Q1: What is Demon Copperhead's role in the story?

E1: Demon Copperhead is the protagonist and narrator of the story, experiencing various challenges and relationships throughout his life.

A1: Protagonist	T1: Role

\vspace{0.5em}

Q2: Who is Demon Copperhead's mother?

E2: Demon Copperhead's mother is a central character in the story, struggling with addiction and neglect.

A2: Addicted mother	T2: Relationship

\vspace{0.5em}

Q3: What is Demon Copperhead's personality like?

E3: Demon Copperhead is portrayed as a resilient and determined character, despite facing numerous hardships.

A3: Resilient and determined	T3: Personality

\vspace{0.5em}

Q4: What event leads to Demon Copperhead's birth?

E4: Demon Copperhead is born to a mother who is struggling with addiction, leading to a difficult start in life.

A4: Difficult birth	T4: Event

\vspace{0.5em}

Q5: Who is Demon Copperhead's foster family?

E5: Demon Copperhead is raised by the Peggots, who become a significant part of his life.

A5: The Peggots	T5: Relationship

\vspace{0.5em}

Q6: What is Demon Copperhead's relationship with Maggot?

E6: Demon Copperhead and Maggot are close friends who grow up together in the same community.

A6: Close friend	T6: Relationship

\vspace{0.5em}

Q7: What is Demon Copperhead's nickname?

E7: Demon Copperhead is given the nickname "Demon" by others, which becomes a part of his identity.

A7: Demon	T7: Personality

\vspace{0.5em}

Q8: What event leads to Demon Copperhead's time with the Peggots?

E8: Demon Copperhead spends time with the Peggots after his mother's struggles with addiction and his own difficult circumstances.

A8: Time with Peggots	T8: Event

\vspace{0.5em}

Q9: What is Demon Copperhead's relationship with Stoner?

E9: Demon Copperhead has a complex relationship with Stoner, who becomes his stepfather and has a significant impact on his life.

A9: Stepfather	T9: Relationship

\vspace{0.5em}

Q10: What event leads to Demon Copperhead's conflict with Stoner?

E10: Demon Copperhead conflicts with Stoner due to Stoner's strict parenting style and his own struggles with identity and self-worth.

A10: Conflict with Stoner	T10: Event

\vspace{0.5em}

\textbf{Reasoning for 2st Chunk:}

\vspace{0.5em}

Q11: What event leads to Demon Copperhead being placed in foster care?

E11: Demon Copperhead's mother becomes addicted to drugs and is involved in a violent relationship, leading to his placement in foster care.

A11: Mother's addiction and abuse.	T11: Event

\vspace{0.5em}

Q12: Who becomes Demon Copperhead's foster father?

E12: Demon Copperhead's foster father is Mr. Crickson, a farmer who takes in foster children.

A12: Mr. Crickson.	T12: Relationship

\vspace{0.5em}

Q13: What event causes Demon Copperhead to run away from home?

E13: Demon Copperhead runs away from home after a violent confrontation with his stepfather, Stoner.

A13: Confrontation with Stoner.	T13: Event

\vspace{0.5em}

Q14: What is Demon Copperhead's attitude towards his foster family?

E14: Demon Copperhead has a complex attitude towards his foster family, initially feeling out of place but gradually forming bonds with some members.

A14: Complex attitude.	T14: Personality

\vspace{0.5em}

Q15: What event leads to Demon Copperhead being placed in a temporary foster home?

E15: Demon Copperhead is placed in a temporary foster home after his mother is hospitalized and unable to care for him.

A15: Mother's hospitalization.	T15: Event

\vspace{0.5em}

[Trace truncated]
}\\
\bottomrule
\end{tabular}
\caption{Example of generated reasoning trace in CroSS dataset. The trace is from the GRPO-trained Qwen-3-8b reasoning model using Lead-128k.}
\label{reasoning_example_1}
\end{table*}

\begin{table*}[ht]
\centering
\begin{tabular}{@{}p{0.7\textwidth} p{0.1\textwidth} p{0.1\textwidth}@{}}
\toprule
\small{}{\textbf{Character:}  John Montraville \quad  \textbf{Book:} Charlotte Temple}\\
\midrule
\multicolumn{3}{@{}p{\textwidth}@{}}{\small{\textbf{Lead-128k:} John Montraville is a young, impetuous soldier who becomes infatuated with Charlotte Temple, a pure and innocent girl, and secretly courts her despite his duty and honor. He is torn between his love for Charlotte and his sense of responsibility, but his passion leads him to elope with her, only to discover she was deceived by Mademoiselle La Rue. Montraville's heart breaks upon learning of Charlotte's betrayal and her subsequent suffering, ultimately leading to his tragic downfall and death, which he brings upon himself in a fit of vengeance against Belcour.

\begin{tabular}{@{}p{0.69\textwidth} >{\centering\arraybackslash}p{0.15\textwidth} >{\centering\arraybackslash}p{0.15\textwidth}@{}}
\small{}{\textbf{Fact}} &  \\
John Montraville is a young soldier. & \textcolor{correct}{\textbf{True}} \\ 
John Montraville is an impetuous soldier. & \textcolor{wrong}{\textbf{False}} \\ 
John Montraville becomes infatuated with Charlotte Temple. & \textcolor{correct}{\textbf{True}} \\
Charlotte Temple is a pure and innocent girl. & \textcolor{wrong}{\textbf{False}} \\ 
John Montraville secretly courts Charlotte Temple despite his duty. & \textcolor{wrong}{\textbf{False}} \\ 
John Montraville secretly courts Charlotte Temple despite his honor. & \textcolor{wrong}{\textbf{False}} \\
John Montraville is torn between his love for Charlotte Temple and his sense of responsibility. & \textcolor{wrong}{\textbf{False}} \\ 
John Montraville's passion leads him to elope with Charlotte Temple. & \textcolor{wrong}{\textbf{False}} \\ 
John Montraville discovers that Charlotte Temple was deceived by Mademoiselle La Rue. & \textcolor{wrong}{\textbf{False}} \\ 
Montraville's heart broke upon learning of Charlotte's betrayal. & \textcolor{wrong}{\textbf{False}} \\ 
Montraville's heart broke upon learning of Charlotte's subsequent suffering. & \textcolor{wrong}{\textbf{False}} \\ 
Montraville's heart breaking led to his tragic downfall. & \textcolor{wrong}{\textbf{False}} \\ 
Montraville's tragic downfall led to his death. & \textcolor{wrong}{\textbf{False}} \\ 
Montraville brought his death upon himself in a fit of vengeance against Belcour. & \textcolor{wrong}{\textbf{False}} \\
\end{tabular}
}}\\
\midrule
\multicolumn{3}{@{}p{\textwidth}@{}}{\small{\textbf{Lead-128k + guided-QA:} John Montraville is a Lieutenant in the army, an ardent and impetuous soldier who becomes infatuated with Charlotte Temple, a young, innocent girl he meets at a boarding school. He secretly courts her, leading her to elope with him, but later discovers her relationship with Belcour, his friend, which causes his heartbreak. Despite his initial honor and compassion, he succumbs to passion, abandons Charlotte, and marries Julia Franklin. His guilt over betraying Charlotte and the death of her child lead to a tragic end, as he kills Belcour in a fit of rage and dies shortly after, haunted by remorse.

\begin{tabular}{@{}p{0.69\textwidth} >{\centering\arraybackslash}p{0.15\textwidth} >{\centering\arraybackslash}p{0.15\textwidth}@{}}
\small{}{\textbf{Fact}} &  \\
John Montraville is a Lieutenant in the army. & \textcolor{wrong}{\textbf{False}} \\ 
John Montraville is an ardent soldier. & \textcolor{wrong}{\textbf{False}} \\ 
John Montraville is an impetuous soldier. & \textcolor{wrong}{\textbf{False}} \\ 
John Montraville became infatuated with Charlotte Temple. & \textcolor{correct}{\textbf{True}} \\ 
Charlotte Temple is a young and innocent girl. & \textcolor{wrong}{\textbf{False}} \\ 
John Montraville met Charlotte Temple at a boarding school. & \textcolor{wrong}{\textbf{False}} \\
John Montraville secretly courted Charlotte Temple. & \textcolor{wrong}{\textbf{False}} \\ 
John Montraville led Charlotte Temple to elope with him. & \textcolor{wrong}{\textbf{False}} \\
John Montraville later discovered Charlotte Temple's relationship with Belcour. & \textcolor{wrong}{\textbf{False}} \\
Belcour is a friend of John Montraville. & \textcolor{correct}{\textbf{True}} \\ 
John Montraville's discovery of Charlotte Temple's relationship with Belcour caused him heartbreak. & \textcolor{wrong}{\textbf{False}} \\
John Montraville initially displayed honor and compassion. & \textcolor{wrong}{\textbf{False}} \\ 
John Montraville succumbed to passion. & \textcolor{correct}{\textbf{True}} \\ 
John Montraville abandoned Charlotte Temple. & \textcolor{correct}{\textbf{True}} \\
John Montraville married Julia Franklin. & \textcolor{correct}{\textbf{True}} \\
John Montraville felt guilt over betraying Charlotte Temple. & \textcolor{correct}{\textbf{True}} \\ 
John Montraville felt guilt over the death of Charlotte Temple's child. & \textcolor{wrong}{\textbf{False}} \\ 
John Montraville killed Belcour in a fit of rage. & \textcolor{wrong}{\textbf{False}} \\ 
John Montraville died shortly after killing Belcour. & \textcolor{wrong}{\textbf{False}} \\
John Montraville was haunted by remorse after his actions. & \textcolor{correct}{\textbf{True}} \\
\end{tabular}
}} \\
\midrule
\multicolumn{3}{@{}p{\textwidth}@{}}{\small{\textbf{Goldstandard Description:} A charming, dashing young officer in the British Army, John Montraville falls in love with Charlotte Temple in Chichester and hopes to marry her. However, his father refuses to endorse a marriage to a poor woman, so Montraville makes Charlotte his mistress and brings her to America with him. He tries to treat her honorably but is easily deceived by his friend Belcour, who ruins his relationship with Charlotte so he can have her as his own mistress. Despite his inability to control his womanizing impulses, Montraville feels guilty about the misfortune he causes Charlotte and others.
}}\\
\bottomrule
\end{tabular}
\caption{Examples of generated descriptions in BookWorm dataset with QA-guided Lead-128k and the corresponding baseline. We demonstrate the facts evaluated based on PRISMA-precision metric.}
\label{tab:eval_examples_2}
\end{table*}

\begin{table*}[ht]
\centering
\begin{tabular}{@{}p{0.7\textwidth} p{0.1\textwidth} p{0.1\textwidth}@{}}
\toprule
\small{}{\textbf{Character:}  John Montraville \quad  \textbf{Book:} Charlotte Temple}\\
\midrule
\multicolumn{3}{@{}p{\textwidth}@{}}{
\small
\textbf{Reasoning for 1st Chunk:}
\vspace{0.5em}

Q1: What role does John Montraville play in the story?

E1: John Montraville is a major character in the story, a young soldier who becomes involved with Charlotte Temple.

A1: Major character	T1: Role

\vspace{0.5em}

Q2: Who is John Montraville related to in the story?

E2: John Montraville is related to Charlotte Temple, as he becomes romantically involved with her.

A2: Charlotte Temple	T2: Relationship

\vspace{0.5em}

Q3: What is John Montraville's personality like?

E3: John Montraville is portrayed as a tender, eloquent, and ardent lover, but also impetuous and driven by his passions.

A3: Ardent and impetuous	T3: Personality

\vspace{0.5em}

Q4: What event leads John Montraville to meet Charlotte Temple?

E4: John Montraville meets Charlotte Temple when he encounters her and her governess at a church in Chichester.

A4: Meeting Charlotte Temple	T4: Event

\vspace{0.5em}

Q5: What decision does John Montraville make regarding his relationship with Charlotte Temple?

E5: John Montraville decides to elope with Charlotte Temple, despite knowing their families would not approve.

A5: Decide to elope	T5: Event

\vspace{0.5em}

Q6: What is John Montraville's profession?

E6: John Montraville is a soldier, as he is mentioned as a Lieutenant in the army.

A6: Soldier	T6: Role

\vspace{0.5em}

Q7: What is John Montraville's relationship with Belcour?

E7: John Montraville is friends with Belcour, who accompanies him on his visits to Charlotte Temple.

A7: Friend	T7: Relationship

\vspace{0.5em}

Q8: What is John Montraville's attitude towards marriage?

E8: John Montraville is initially cautious about marriage, as his father warns him against rushing into it.

A8: Cautious	T8: Personality

\vspace{0.5em}

Q9: What is one of the events involving John Montraville and Charlotte Temple?

E9: One event involving John Montraville and Charlotte Temple is their clandestine meetings at a summer-house.

A9: Clandestine meetings	T9: Event

\vspace{0.5em}

Q10: What is John Montraville's relationship with his father?

E10: John Montraville has a respectful relationship with his father, who gives him advice about marriage and duty.

A10: Respectful	T10: Relationship

\vspace{0.5em}

\textbf{Reasoning for 2st Chunk:
}
\vspace{0.5em}

Q11: Who is John Montraville romantically involved with?

E11: John Montraville is romantically involved with Charlotte Temple and Julia Franklin.

A11: Charlotte Temple	T11: Relationship

\vspace{0.5em}

Q12: What event leads to John Montraville's conflict with Charlotte Temple?

E12: John Montraville's conflict with Charlotte Temple arises from her elopement and his subsequent discovery of her relationship with Belcour.

A12: Elopement	T12: Event

\vspace{0.5em}

Q13: What event leads to John Montraville's conflict with Julia Franklin?

E13: John Montraville's conflict with Julia Franklin arises from his indecision about marrying her while still being involved with Charlotte Temple.

A13: Indecision	T13: Event

\vspace{0.5em}

Q14: What event leads to John Montraville's discovery of Charlotte's infidelity?

E14: John Montraville discovers Charlotte's infidelity when he finds her in bed with Belcour.

A14: Discovery	T14: Event

\vspace{0.5em}

Q15: What is John Montraville's role in the story's resolution?

E15: John Montraville plays a significant role in the story's resolution by ultimately leaving Charlotte and pursuing a relationship with Julia Franklin.

A15: Pursuing relationship	T15: Event

\vspace{0.5em}

Q16: What is John Montraville's personality trait regarding honor?

E16: John Montraville is portrayed as having a strong sense of honor, which conflicts with his romantic entanglements.

A16: Strong sense of honor	T16: Personality

\vspace{0.5em}
[Trace truncated]
}\\
\bottomrule
\end{tabular}
\caption{Example of generated reasoning trace in BookWorm dataset. The trace is from the GRPO-trained Qwen-3-8b reasoning model with Lead-128k as input context.}
\label{reasoning_example_2}
\end{table*}


\end{document}